\documentclass[sigconf]{acmart}

\usepackage{graphicx,multirow,paralist,url,xspace,xcolor,nicefrac,subfigure}
\usepackage{amsmath,amsthm,amsfonts,amsfonts}
\usepackage[ruled,vlined,linesnumbered]{algorithm2e}
\usepackage{balance}

\usepackage{pifont,color}

\newcommand*{\scale}[2][4]{\scalebox{#1}{$#2$}}

\AtBeginDocument{%
  \providecommand\BibTeX{{%
    \normalfont B\kern-0.5em{\scshape i\kern-0.25em b}\kern-0.8em\TeX}}}

\begin{document}

\title{Cross-Network Learning with Partially Aligned Graph Convolutional Networks}
\author{Meng Jiang}
\affiliation{
  \institution{Department of Computer Science and Engineering, University of Notre Dame}
  \streetaddress{384 Fitzpatrick Hall}
  \city{Notre Dame}
  \state{Indiana}
  \country{USA}}
\email{mjiang2@nd.edu}


\begin{abstract}
Graph neural networks have been widely used for learning representations of \emph{nodes} for many downstream tasks on graph data.
Existing models were designed for the nodes on a single graph, which would not be able to utilize information across multiple graphs.
The real world does have multiple graphs where the nodes are often \emph{partially aligned}. For examples, knowledge graphs share a number of named entities though they may have different relation schema; collaboration networks on publications and awarded projects share some researcher nodes who are authors and investigators, respectively; people use multiple web services, shopping, tweeting, rating movies, and some may register the same email account across the platforms.
In this paper, I propose partially aligned graph convolutional networks to learn node representations across the models.
I investigate multiple methods (including model sharing, regularization, and alignment reconstruction) as well as theoretical analysis to \emph{positively} transfer knowledge across the (small) set of partially aligned nodes.
Extensive experiments on real-world knowledge graphs and collaboration networks show the superior performance of our proposed methods on relation classification and link prediction.

\end{abstract}

\maketitle

\section{Introduction}
\label{sec:intro}
\begin{figure}[t]
    \centering
    \includegraphics[width=0.46\textwidth]{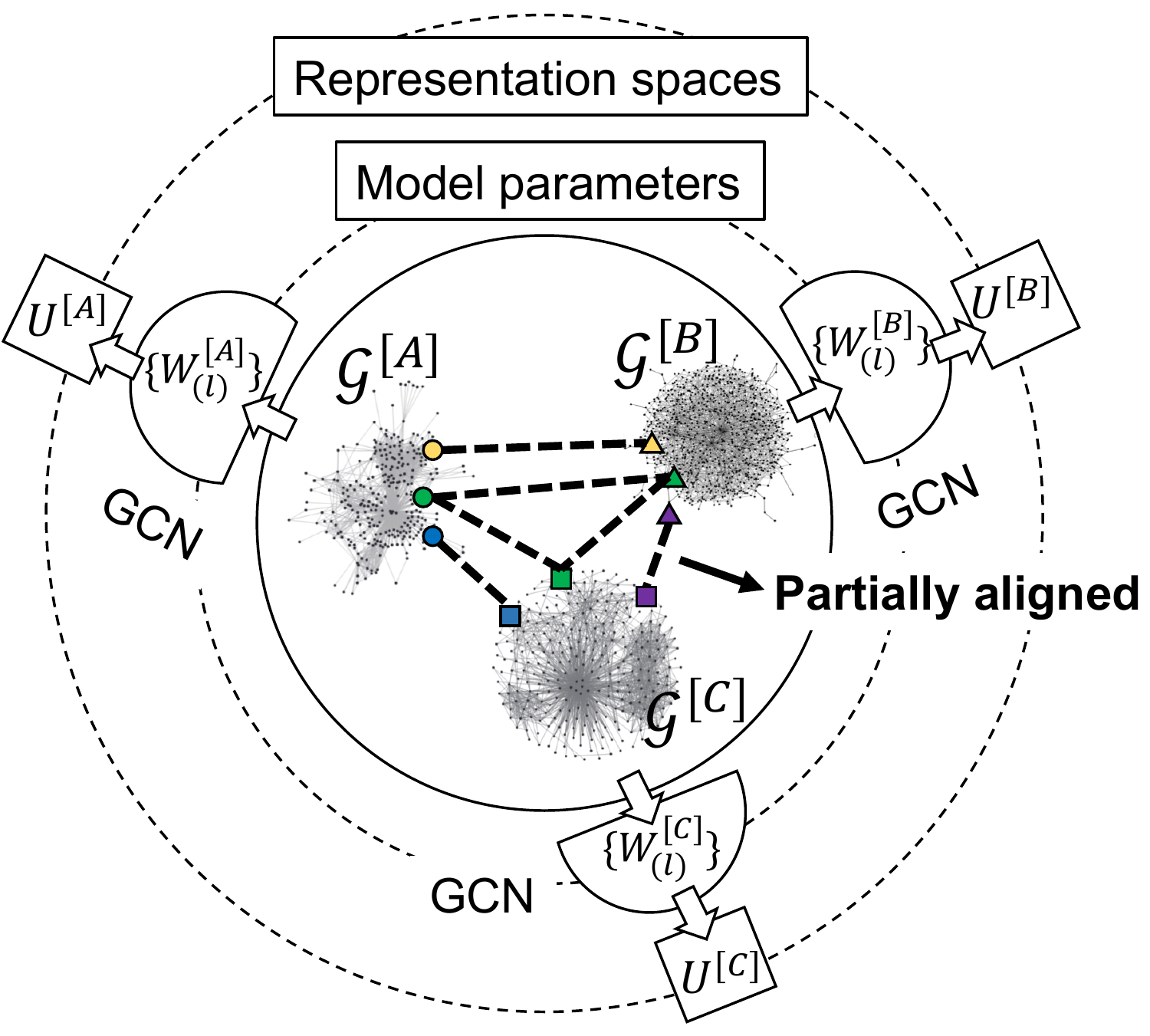}
    \caption{Cross-GCN learning uses network alignment information to bridge the separated model parameters and representation spaces and transfer knowledge across the models.}
    \label{fig:idea}
\end{figure}

Graph learning represents or encodes graph structure to perform tasks such as classification, clustering, and regression on graphs.
It is important and ubiquitous with applications ranging from knowledge graph completion to movie recommendation.
It is originally studied as dimensionality reduction in adjacency matrix factorization methods \cite{cai2010graph}.
Then network embedding methods aim at preserving particular structure, property, or side information in graph or networked data \cite{cui2018survey,dong2017metapath2vec}.
Recently, graph convolutional networks (GCNs) have been introduced to specify model parameters (e.g., a set of weight matrices) on how to aggregate information from a node's local neighborhood \cite{kipf2017semi,hamilton2017representation}.
Unlike the network embedding methods, these parameters are shared across nodes and allow the approaches to generate node representations that do not necessarily rely on the entire graph \cite{velivckovic2017graph,hamilton2017inductive,wang2019heterogeneous}.

The sparsity problem in graph data is a major bottleneck for most graph learning methods.
For example, knowledge graphs are incomplete, causing inaccuracy in downstream tasks such as fact retrieval and question answering, especially when knowledge is represented under the Open World Assumption \cite{schlichtkrull2018modeling}.
In real-world recommender systems, users can rate a very limited number of items, so the user-item bipartite graph is always extremely sparse \cite{ying2018graph,fan2019graph}.
The observed data that can be used for adjacency matrix factorization or neural net graph learning are radially insufficient.

\begin{figure*}[t]
    \centering
    \includegraphics[width=\textwidth]{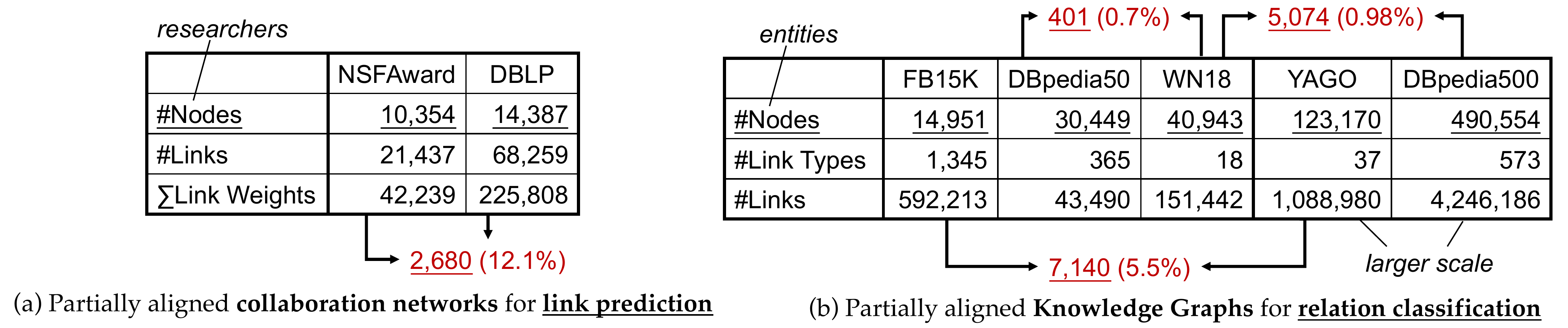}
    \caption{Examples of partially aligned networks in real world for two different tasks.}
    \label{fig:datasets}
\end{figure*}

Although we cannot fabricate more observations, we may borrow useful knowledge from graph data on different sources, domains, or platforms.
Consider the following case:
A new UberPool-like carpooling App has launched.
Due to lack of usage at the beginning, machine learning on the very sparse user-user ride-sharing graph cannot be effective for ride recommendation.
Now suppose that a small number of the users register the App with their social media accounts, and we have social graph data available from the social media platform.
Then we ask, though the percentage of the ``aligned'' users is very small (10\%, 5\%, or even less than 1\%), can we use the alignment as a bridge and transfer knowledge from the social graph to the ride-sharing graph?
Since ride sharing and social relationship are somewhat related, the \emph{partially} aligned users on different graphs (platforms) can share similar behavioral patterns.
Thus, learning across the graphs can be beneficial.
When the graph learning models are state-of-the-art, learning across graph convolutional networks can transfer knowledge via the partially aligned nodes.

Figure~\ref{fig:idea} illustrates the idea of cross-GCN learning. Originally, the three GCN models are trained \emph{separately} to generate low-dimensional representations of nodes $U^{[\cdot]}$ with a set of model parameters $\{W^{[\cdot]}_{(l)}\}$. We seek for effective methods to transfer knowledge across the GCN models through the partial alignment information between $\mathcal{G}^{[\cdot]}$ into their node representations.

In this paper, I study how to alleviate the sparsity problem in graph learning by transferring knowledge across multiple GCN models, when the graphs are partially aligned.

The contributions of this work can be described in three aspects:
\begin{enumerate}
\item \textbf{Cross-GCN learning methods:}
I design three different types of methods: (a) sharing model parameters, (b) aligning representation spaces with regularizations, and (c) simultaneously reconstructing the alignment and learning representations. I present how to generalize them for more than two GCNs and for relational graphs.
\item \textbf{Theoretical foundations:}
I present theories on (a) the choice of the number of dimensions of the representation spaces to force knowledge transfer and (b) the conditions of positive transfer across GCN models, regularizations, and alignment reconstruction.
\item \textbf{Extensive experiments:}
As shown in Figure~\ref{fig:datasets}, I perform experiments on two types of graphs datasets and graph learning tasks: (a) link prediction on weighted graphs (e.g., collaboration social networks) and (b) relation classification on knowledge graphs. Both graphs are partially aligned (see the number of aligned nodes in red).
Cross-GCN learning performs consistently better than individual GCNs on all the datasets and tasks.
\end{enumerate}

The rest of this paper is organized as follows. Section~\label{sec:prelim} gives definitions of basic concepts (e.g., graph convolutional network) and the proposed goal. Section~\label{sec:methods} presents multiple kinds of methods to achieve the goal of cross-network learning. Section~\label{sec:theory} provides theoretical analysis on choice of number of dimensions and positive transfer. Section ~\label{sec:experiments} presents experimental results on knowledge graphs and collaboration networks for relation classification and link prediction, respectively. Section~\label{sec:related} reviews the related papers, and Section~\label{sec:conclusions} concludes the work.

\section{Preliminaries}
\label{sec:prelim}
\subsection{Graphs and Learning Tasks}

We study on two types of graphs and related tasks about learning, classifying, or predicting links between nodes.

\begin{definition}[Weighted graph] A weighted graph, denoted as $\mathcal{G}=(\mathcal{U},\mathcal{E}\subset\mathcal{U}\times\mathcal{U})$, contains a set of nodes $\mathcal{U}$ and a set of links $\mathcal{E}$, with a weight mapping function $w:\mathcal{E}\rightarrow\mathbb{R}$.
\end{definition}

For example, collaboration networks are weighted graphs. The weight is the frequency of collaborations between two person nodes. So, it is a positive integer when the link exists. The task of \emph{link prediction} is to predict whether the value of $w(u_i, u_j)$ is positive or what the value is, given a pair of nodes $u_i, u_j \in \mathcal{U}$.

\begin{definition}[Relational graph] A relational graph, denoted as $\mathcal{G}_{rel}=(\mathcal{U},\mathcal{E}\subset\mathcal{U}\times\mathcal{U}\times\mathcal{R}, \mathcal{R})$, contains a set of nodes $\mathcal{U}$, a set of relational links $\mathcal{E}$, and a relation schema $\mathcal{R}$. A relation link $e=(u_i,u_j,r)\in\mathcal{E}$ indicates that $u_i$ and $u_j$ have relation $r$.
\end{definition}

Knowledge graphs are typical relational graphs, where the relational links need to be learned and completed. The task of \emph{relation classification} is to predict the type of relation between two nodes (i.e., entities) $u_i, u_j$ from the relation schema $\mathcal{R}$ so that a link $e=(u_i,u_j,r)$ is likely to exist in the graph.



The fundamental problem of both tasks is \emph{node representation learning} that learns low-dimensional vectors of nodes to preserve the graph's information. Take \emph{link prediction} as an example -- formally, it is to learn a mapping function: $\vec{u} = f(u, \mathcal{G}, \Theta) \in \mathbb{R}^d$, where $\Theta$ is the node representation learning model's parameters and $d$ is the number of dimensions of node representations. Then the value of $w(u_i, u_j)$ is assigned as $\sigma(\vec{u_i}^{\top} \vec{u_j})$, where $\sigma$ is sigmoid. Simple predictive functions can be specified for the other two tasks.

\subsection{Graph Convolutional Network}

Graph convolutional network (GCN) is one of the most popular models for node representation learning \cite{kipf2017semi}. Let us continue using \emph{link prediction} as the example. We denote the number of nodes in graph $\mathcal{G}$ by $n = |\mathcal{U}|$. Suppose $A \in \mathbb{R}^{n \times n}$ is the adjacency matrix of $\mathcal{G}$ and $X \in \mathbb{R}^{n \times m}$ has the nodes' raw attribute values, where $m$ is the number of the raw attributes. A two-layer GCN generates the matrix of node representations as follows:
\begin{equation}
    U = \mathrm{softmax} \left( \hat{A}~\mathrm{ReLU}(\hat{A}~X~W_{(1)})~W_{(2)} \right) \in \mathbb{R}^{n \times d},
    \label{eq:gcn}
\end{equation}
where $\Theta = \{W_{(1)} \in \mathbb{R}^{m \times d}, W_{(2)} \in \mathbb{R}^{d \times d}\}$ has layer-specific weight matrices. We choose two layers as most of existing studies suggest. The pre-processing step includes (1) $\tilde{A} = A+I_n$, (2) $\tilde{D}_{ii} = \sum_{j} \tilde{A}_{ij}$, and (3) $\hat{A} = \tilde{D}^{\frac{1}{2}}~\tilde{A}~\tilde{D}^{\frac{1}{2}}$. Graph reconstruction is used to inform the GCN to learn $\Theta$ to generate effective $U$:
\begin{equation}
    g(X, A, W_{(1)}, W_{(2)}) = \sigma(U U^{\top}) \in {[0,1]}^{n \times n}.
\end{equation}
The loss function can be written as follows:
\begin{eqnarray}
    f(W_{(1)}, W_{(2)}) & = & \mathcal{L} \left( g(X, A, W_{(1)}, W_{(2)}),~A \right) \nonumber \\
    & = & - \sum_{i,j \in \{1 \dots n\}} A_{ij}~\mathrm{log} \left( \sigma(\vec{u_i}^{\top} \vec{u_j}) \right)
    \label{eq:lossgcn}
\end{eqnarray}
where $\vec{u_i}$ and $\vec{u_j}$ are the $i$-th and $j$-th rows in matrix $U$. It is worth noting that one can easily extend our study from two-layer GCN to $l$-layer GCN ($l > 2$).

\subsection{Cross-Network Learning}

One GCN can be built for one graph. What if we have multiple graphs which overlap each other to perform the same type of tasks? The overlapping parts, referred as \emph{network alignment} serve as bridge for transferring knowledge across multiple GCNs so that they can mutually enhance each other. Without loss of generality, we focus on the cases that have two graphs $\mathcal{G}^{[A]}$ and $\mathcal{G}^{[B]}$. We denote by $A^{[A,B]} \in {\{0,1\}}^{n_A \times n_B}$ the adjacency matrix of network alignment between the two graphs: $A^{[A,B]}_{ij} = 1$, if node $u^{[A]}_i$ and node $u^{[B]}_j$ are aligned; where $n_A = |\mathcal{U}^{[A]}|$ and $n_B = |\mathcal{U}^{[B]}|$. Usually the graphs are \emph{partially aligned}, leading to missing alignment links and the sparsity of the alignment matrix.

\begin{definition}[Separated GCNs] Traditionally, the network alignment was ignored in building GCNs. So, two separated GCNs would be built for the tasks on graphs $\mathcal{G}^{[A]}$ and $\mathcal{G}^{[B]}$:
\begin{eqnarray}
    f(W^{[A]}_{(1)}, W^{[A]}_{(2)}) = \mathcal{L} \left( g(X^{[A]}, A^{[A]}, W^{[A]}_{(1)}, W^{[A]}_{(2)}),~A^{[A]} \right), \nonumber \\
    f(W^{[B]}_{(1)}, W^{[B]}_{(2)}) = \mathcal{L} \left( g(X^{[B]}, A^{[B]}, W^{[B]}_{(1)}, W^{[B]}_{(2)}),~A^{[B]} \right),
\end{eqnarray}
where no model parameters (between ${\Theta}^{[A]}$ and ${\Theta}^{[B]}$) is shared and no alignment information is used.
\end{definition}

\begin{definition}[Cross-network learning] Given two graphs $\mathcal{G}^{[A]}$ and $\mathcal{G}^{[B]}$ as well as the network alignment information $A^{[A,B]}$, build cross-network learning models that have the properties below:
\begin{itemize}
    \item generate node representations $U^{[A]} \in \mathbb{R}^{n_A \times d_A}$ and $U^{[B]} \in \mathbb{R}^{n_B \times d_B}$ through graph convolutions, where $d_A$ and $d_B$ are the numbers of dimensions, respectively -- $d_A \ll m_A$, $d_B \ll m_B$, and they do not have to be equivalent;
    \item perform better than separated GCNs by sharing model parameters and/or incorporating the network alignment information.
\end{itemize}
\end{definition}

\section{Proposed Methods}
\label{sec:methods}
In this section, we present multiple types of methods for transferring knowledge across GCNs. The first type trains two GCNs by sharing their model parameters (Sections~\ref{sec:sharingmodel} and~\ref{sec:trainingstrategy}). The second uses the network alignment information to regularize the output representation spaces (Section~\ref{sec:aligningspaces}). And the third type uses network alignment reconstruction as an extra task for knowledge transfer (Section~\ref{sec:alignmentreconstruction}). Generalizations to multiple GCNs and particular types of graphs are discussed (Sections~\ref{sec:multigcn} to~\ref{sec:bipartite}).

\subsection{Sharing Model Parameters across GCNs}
\label{sec:sharingmodel}

Recall that a two-layer GCN on graph $\mathcal{G}^{[A]}$ has parameters $\Theta^{[A]}=\{W^{[A]}_{1} \in \mathbb{R}^{m_A \times d_A}, W^{[A]}_{2} \in \mathbb{R}^{d_A \times d_A}\}$. Same goes for graph $\mathcal{G}^{[B]}$.

Suppose $d_A \neq d_B$. None of the parameters can be directly shared.

Suppose $d_A = d_B$. The second-layer weight parameters can be shared: $W_{2} = W^{[A]}_{2} = W^{[B]}_{2}$. So, we can perform joint GCN training:
\begin{eqnarray}
    \scale[0.9]{f(W^{[A]}_{(1)}, W^{[B]}_{(1)}; W_{(2)}) = \alpha^{[A]} \cdot \mathcal{L} \left( g(X^{[A]}, A^{[A]}, W^{[A]}_{(1)}, W_{(2)}),~A^{[A]} \right)} \nonumber \\
    \scale[0.9]{+~\alpha^{[B]} \cdot \mathcal{L} \left( g(X^{[B]}, A^{[B]}, W^{[B]}_{(1)}, W_{(2)}),~A^{[B]} \right),}
    \label{eq:sharew2}
\end{eqnarray}
where $\alpha^{[A]}$ and $\alpha^{[B]}$ are per-graph weight of the training procedure. When we have two graphs, $\alpha^{[A]} + \alpha^{[B]} = 1$.

Suppose $m_A = m_B$ and $d_A = d_B$. The weight parameters of both layers can be shared across GCNs. We omit the equation for space -- it simply replaces $W^{[A]}_{(1)}$ and $W^{[B]}_{(1)}$ in Eq.(\ref{eq:sharew2}) by $W_{(1)}$. However, the multiple graphs can hardly have the same raw attribute space. So, when $d = d_A = d_B$ but $m_A \neq m_B$, we assume that linear factorized components of $W^{[A]}_{(1)}$ and $W^{[B]}_{(1)}$ can be shared. We denote the shared component by $W_{(1)} \in \mathbb{R}^{\hat{m} \times d}$, where $\hat{m} = \mathrm{min}\{m_A, m_B\}$ or a smaller value. Then the loss function of joint training is as follows:
\begin{eqnarray}
    \scale[0.8]{f(Q^{[A]}, Q^{[B]}; W_{(1)}, W_{(2)}) = \alpha^{[A]} \cdot \mathcal{L} \left( g(X^{[A]}Q^{[A]}, A^{[A]}, W_{(1)}, W_{(2)}),~A^{[A]} \right)} \nonumber \\
    \scale[0.8]{+~\alpha^{[B]} \cdot \mathcal{L} \left( g(X^{[B]}Q^{[B]}, A^{[B]}, W_{(1)}, W_{(2)}),~A^{[B]} \right),}
    \label{eq:sharew1}
\end{eqnarray}
where $Q^{[A]} \in \mathbb{R}^{m_A \times \hat{m}}$ and $Q^{[B]} \in \mathbb{R}^{m_B \times \hat{m}}$ are linear transformation matrices on raw feature spaces for aligning the first-layer weight parameter matrices.

When the number of graphs is two, we can perform alternatives to save one linear transformation matrix. Alternative 1 is to use $Q^{[A \rightarrow B]} \in \mathcal{R}^{m_A \times m_B}$ to align $\mathcal{G}^{[A]}$'s raw attribute space to $\mathcal{G}^{[B]}$'s:
\begin{eqnarray}
    \scale[0.8]{f(Q^{[A \rightarrow B]}; W^{[B]}_{(1)}, W_{(2)}) = \alpha^{[A]} \cdot \mathcal{L} \left( g(X^{[A]}Q^{[A \rightarrow B]}, A^{[A]}, W^{[B]}_{(1)}, W_{(2)}),~A^{[A]} \right)} \nonumber \\
    \scale[0.8]{+~\alpha^{[B]} \cdot \mathcal{L} \left( g(X^{[B]}, A^{[B]}, W^{[B]}_{(1)}, W_{(2)}),~A^{[B]} \right),}
\end{eqnarray}
And alternative 2 is to use $Q^{[B \rightarrow A]} \in \mathcal{R}^{m_B \times m_A}$ to align $\mathcal{G}^{[B]}$'s raw attribute space to $\mathcal{G}^{[A]}$'s.

\subsection{Training Strategies across GCNs}
\label{sec:trainingstrategy}

The training strategy presented in the above section is \emph{joint training} or referred as multi-task learning. The weights of the tasks, i.e., $\alpha^{[A]}$ and $\alpha^{[B]}$, are determined prior to the training procedure. The other popular training strategy is called \emph{pre-training}, that is to define one task a time as the target task and all the other tasks as source tasks. Take the method of sharing $W_{(2)}$ as an example. The loss function of joint training was given in Eq.(\ref{eq:sharew2}). Now if learning graph $\mathcal{G}^{[B]}$ is the target task, then learning graph $\mathcal{G}^{[A]}$ is the source task. The pre-training procedure has two steps. Their loss functions are:
\begin{eqnarray}
    f_1(W^{[A]}_{(1)}; W_{(2)}) = \mathcal{L} \left( g(X^{[A]}, A^{[A]}, W^{[A]}_{(1)}, W_{(2)}),~A^{[A]} \right), \nonumber \\
    f_2(W^{[B]}_{(1)}; W_{(2)}) = \mathcal{L} \left( g(X^{[B]}, A^{[B]}, W^{[B]}_{(1)}, W_{(2)}),~A^{[B]} \right).
\end{eqnarray}
Each step still minimizes Eq.(\ref{eq:sharew2}): specifically, the first step uses $\alpha^{[A]} = 1$ and $\alpha^{[B]} = 0$; and the second step uses $\alpha^{[A]} = 0$ and $\alpha^{[B]} = 1$. The first step warms up the training of the shared $W_{(2)}$ on $\mathcal{G}^{[A]}$ so that the ``fine-tuning'' step on $\mathcal{G}^{[B]}$ can find more effective model parameters than starting from pure randomization. When the source and target are swapped, the two steps are swapped.

\subsection{Aligning Representation Spaces with Regularizations}
\label{sec:aligningspaces}

Besides sharing model parameters, an idea of bridging the gap between two GCNs is to align their output representation spaces. Given GCNs trained on two graphs $\mathcal{G}^{[A]}$ and $\mathcal{G}^{[B]}$, if two nodes $u^{[A]}_i$ and $u^{[B]}_j$ are aligned, we assume that their representations, $\vec{u}^{[A]}_i \in \mathbb{R}^{d_A}$ and $\vec{u}^{[B]}_j \in \mathbb{R}^{d_B}$ are highly correlated. For example, one named entity (person, location, or organization) in two different knowledge graphs should have very similar representations. The same goes for one researcher in two collaboration networks (as an investigators in projects or an author in papers).

We add the network alignment information as two types of regularization terms into the loss function. The first type is \emph{Hard regularization}. It assumes that the aligned nodes have exactly the same representations in the two output spaces. This requires the two spaces to have the same number of dimensions: $d_A = d_B$. The term is written as:
\begin{equation}
    h(W^{[A]}_{(1)}, W^{[A]}_{(2)}, W^{[B]}_{(1)}, W^{[B]}_{(2)}) = {\left\| U^{[A]} - A^{[A,B]}~U^{[B]} \right\|}^2_\mathrm{F}.
\end{equation}
The entire loss function is as below:
\begin{eqnarray}
    & & f({\Theta}^{[A]}, {\Theta}^{[B]}) = f(W^{[A]}_{(1)}, W^{[A]}_{(2)}, W^{[B]}_{(1)}, W^{[B]}_{(2)}) \nonumber \\
    & = & (1-\beta) \cdot \alpha^{[A]} \cdot \mathcal{L} \left( g(X^{[A]}, A^{[A]}, W^{[A]}_{(1)}, W^{[A]}_{(2)}),~A^{[A]} \right) \nonumber \\
    & & +~(1-\beta) \cdot  \alpha^{[B]} \cdot \mathcal{L} \left( g(X^{[B]}, A^{[B]}, W^{[B]}_{(1)}, W^{[B]}_{(2)}),~A^{[B]} \right) \nonumber \\
    & & +~\beta \cdot h(W^{[A]}_{(1)}, W^{[A]}_{(2)}, W^{[B]}_{(1)}, W^{[B]}_{(2)}),
    \label{eq:alignreg}
\end{eqnarray}
where $\beta$ is the weight of the regularization task.

The second type is \emph{Soft regularization} which is designed for more common cases that $d_A \neq d_B$. It assumes that the aligned nodes' representations from one GCN can be linearly transformed into the ones in the other GCN's output space:
\begin{equation}
    h({\Theta}^{[A]}, {\Theta}^{[B]}, R^{[A \rightarrow B]}) = {\left\| U^{[A]} R^{[A \rightarrow B]} - A^{[A,B]}~U^{[B]} \right\|}^2_\mathrm{F},
    \label{eq:softreg}
\end{equation}
where $R^{[A \rightarrow B]} \in \mathbb{R}^{d_A \times d_B}$ is the transformation matrix. Note that $R^{[A \rightarrow B]}$ is treated as model parameters in the loss function.

\subsection{Alignment Reconstruction}
\label{sec:alignmentreconstruction}

When the two graphs are partially aligned, the network alignment information is usually sparse and incomplete. To address this issue, we treat the alignment information as a bipartite graph $\mathcal{G}^{[A,B]}$ and learn to reconstruct and complete the graph. Given the node representations $U^{[A]} \in \mathbb{R}^{n_A \times d_A}$ and $U^{[B]} \in \mathbb{R}^{n_B \times d_B}$, can we reconstruct the adjacency matrix of observed alignment data $\mathcal{G}^{[A,B]}$, i.e., $I^{[A,B]} \odot A^{[A,B]} \in \mathcal{R}^{n_A \times n_B}$, where $I^{[A,B]}$ is the indicator matrix of the observations? We introduce $R^{[A,B]} \in \mathbb{R}^{d_A \times d_B}$ and minimize the following optimization term:
\begin{equation}
    h = {\left\| \sigma \left( U^{[A]}~R^{[A,B]}~U^{[B]\top} \right) - I^{[A,B]} \odot A^{[A,B]} \right\|}^2_\mathrm{F}.
    \label{eq:alignreconstruction}
\end{equation}
The entire loss function optimizes on three graph reconstruction tasks ($\mathcal{G}^{[A]}$, $\mathcal{G}^{[B]}$, and $\mathcal{G}^{[A,B]}$), by replacing $h$ in Eq.(\ref{eq:alignreg}) with the above equation. The learning procedure transfers knowledge across the three tasks to find effective representations $U^{[A]}$ and $U^{[B]}$.

\subsection{Beyond Two: Extend to Multiple GCNs}
\label{sec:multigcn}

Suppose we have $K > 2$ GCNs. All the above methods (i.e., loss functions) can be easily extended from $2$ to $K$ GCNs. For example, the loss of joint multi-GCN training with \emph{shared $W_{2}$} can be written as below (extended from Eq.(\ref{eq:sharew2})):
\begin{equation}
    \scale[0.9]{f({\{W^{[i]}_{(1)}\}}|^{K}_{i=1}; W_{(2)}) = \sum^{K}_{i=1} \alpha^{[i]} \mathcal{L} \left( g(X^{[i]}, A^{[i]}, W^{[i]}_{(1)}, W_{(2)}),~A^{[i]} \right).}
\end{equation}
And the loss of joint multi-GCN training with \emph{shared $W_{1}$ and $W_{2}$} can be written as below (extended from Eq.(\ref{eq:sharew1})):
\begin{equation}
    \scale[0.88]{f({\{Q^{[i]}\}}|^{K}_{i=1}; W_{(1)}, W_{(2)}) = \sum^{K}_{i=1} \alpha^{[i]} \mathcal{L} \left( g(X^{[i]}Q^{[i]}, A^{[i]}, W_{(1)}, W_{(2)}),~A^{[i]} \right).}
\end{equation}

When using the network alignment information, the \emph{soft regularization} term is written as below (extended from Eq.(\ref{eq:softreg})):
\begin{eqnarray}
    & & h \left( {\{{\Theta}^{[i]}\}}|^{K}_{i=1},~{\{R^{[i \rightarrow j]}\}}|_{i < j;~i, j \in \{1 \dots n\}} \right) \nonumber \\
    & = & \sum_{i < j;~i, j \in \{1 \dots n\}} \gamma^{[i \rightarrow j]} {\left\| U^{[i]} R^{[i \rightarrow j]} - A^{[i,j]}~U^{[j]} \right\|}^2_\mathrm{F},
\end{eqnarray}
where $\gamma^{[i \rightarrow j]}$ is the regularization weight of aligning representation spaces between graphs $\mathcal{G}^{[i]}$ and $\mathcal{G}^{[j]}$.

The optimization term for alignment reconstruction and completion can be written as below (extended from Eq.(\ref{eq:alignreconstruction})):
\begin{equation}
    h = \sum_{i < j} \gamma^{[i \rightarrow j]} {\left\| \sigma \left( U^{[i]}~R^{[i,j]}~U^{[j]\top} \right) - I^{[i,j]} \odot A^{[i,j]} \right\|}^2_\mathrm{F}.
    \label{eq:alignreconstruction}
\end{equation}

\subsection{Implementation on Relational Graphs: Relation Classification across GCNs}
\label{sec:relational}

Schlichtkrull \emph{et al.} proposed neural relational modeling that uses GCNs for relation classification on relational graphs \cite{schlichtkrull2018modeling}. The output node representations from two-layer GCN are given as below:
\begin{equation}
    U = \mathrm{softmax} \left( \sum_{r \in \mathcal{R}} \frac{1}{c_r} A_{:,:,r}~\mathrm{ReLU} \left( \sum_{r \in \mathcal{R}} \frac{1}{c_r} A_{:,:,r}~X~W_{r,(1)} \right)~W_{r,(2)} \right).
\end{equation}
Clearly, it can be derived from Eq.(\ref{eq:gcn}) by making these changes:
\begin{itemize}
    \item The relational graph $\mathcal{G}_{rel} = \{\mathcal{U}, \mathcal{E}, \mathcal{R}\}$ is denoted as a three-way tensor $A \in {\{0,1\}}^{n \times n \times n_R}$, where $n = |\mathcal{U}|$ is the number of entity nodes and $n_R = |\mathcal{R}|$ is the number of relation types;
    \item The model parameters are two three-way tensors: \\ $\Theta = \{{W_{r,(1)}}|_{r \in \mathcal{R}} \in \mathbb{R}^{m \times d}, {W_{r,(2)}}|_{r \in \mathcal{R}} \in \mathbb{R}^{d \times d}\}$;
    \item $c_r$ controls the weight of relation type $r$.
\end{itemize}

Relation classification can be considered as relational link prediction if we apply the prediction on all the possible entity-relation triples. The full prediction matrix is:
\begin{equation}
    g(X, A, W_{(1)}, W_{(2)}, D) = \sigma(U \otimes D \otimes U^{\top}) \in {[0,1]}^{n \times n \times n_R},
\end{equation}
where $D \in \mathbb{R}^{d \times d \times n_R}$ is a three-way core tensor. For each relation type $r$, $D_{:,:,r}$ is a $d \times d$-diagonal matrix. The loss function is below:
\begin{eqnarray}
    & & f(W_{(1)}, W_{(2)}, D) = \mathcal{L} \left( g(X, A, W_{(1)}, W_{(2)}, D),~A \right) \nonumber \\
    & = & - \sum_{i,j \in \{1 \dots n\};~r \in \mathcal{R}} A_{i,j
    ,r}~\mathrm{log} \left( \sigma(\vec{u_i}^{\top}~D_{:,:,r}~\vec{u_j}) \right)
\end{eqnarray}
All the proposed methods for partially aligned GCNs in the previous sections can be applied here. The key differences are (1) a three-way tensor data for reconstruction and (2) additional parameters $D$.

\section{Theoretical Analysis}
\label{sec:theory}
In this section, we present theoretical results for partially aligned GCNs. First, we study the choices of the number of dimensions of node representations that may perform no knowledge transfer across GCNs. Then, we study two factors of positive transfer across GCNs. Note that many details are given in Appendix to save space.

\subsection{Choices of the Number of Dimensions $d$}

The task of graph reconstruction is to use the node representations $U \in \mathbb{R}^{n \times d}$ to recover the adjacency matrix $A \in \mathbb{R}^{n \times n}$. With singular value decomposition, we have $A \simeq \hat{U}~\Sigma~\hat{U}^{\top}$, where $\hat{U} \in \mathbb{R}^{n \times \mathrm{rank}(A)}$ and $\Sigma$ is square diagonal of size $\mathrm{rank}(A) \times \mathrm{rank}(A)$. So, the loss function in Eq.(\ref{eq:lossgcn}) can be approximated as the squared loss:
\begin{equation}
    f = {\left\|~U~B - \hat{U}~\sqrt{\Sigma}~\right\|}^2_\mathrm{F},
\end{equation}
where $B \in \mathcal{R}^{d \times \mathrm{rank}(A)}$. It transforms the $n \times n$ classification tasks (for link prediction) into $\mathrm{rank}(A)$ regression tasks.

\begin{proposition}
Suppose we have $K$ GCNs. In a linear setting for every model in the partially aligned GCNs, the loss function becomes:
\begin{equation}
    \scale[0.95]{f \left( {\{\Theta_{i} = W^{[i]}_{(1)} W^{[i]}_{(2)} B^{[i]}\}}|^{K}_{i=1} \right) = \sum^{K}_{i=1} { {\left\|~\hat{A}^{[i]~2} X^{[i]} \Theta_{i}  - \hat{U}^{[i]}~\sqrt{{\Sigma}^{[i]}}~\right\|}^2_\mathrm{F} }.}
    \label{eq:losslinear}
\end{equation}
The optimal solution for the $i$-th GCN is
\begin{equation}
    {\Theta}_{i} = {\left( X^{[i]\top} \hat{A}^{[i]~4} X^{[i]} \right)}^{\dagger} X^{[i]\top} \hat{A}^{[i]~2} \hat{U}^{[i]} \sqrt{{\Sigma}^{[i]}}  \in \mathcal{R}^{m_i \times \mathrm{rank}(A)},
\end{equation}
where $X^{\dagger}$ denotes $X$'s pseudoinverse. Hence a capacity of $\mathrm{rank}(A^{[i]})$ suffices for the $i$-th GCN. If $d \geq \sum^{K}_{i=1} \mathrm{rank}(A^{[i]})$, then $B^{[i]}$ can be an identity matrix and there is \emph{no transfer} between any two GCNs. In that case, no matter ${\{W^{[i]}_{(1)}\}}|^{K}_{i=1}$ or ${\{W^{[i]}_{(2)}\}}|^{K}_{i=1}$ are shared or not, these exists an optimum $W^{[i]*}_{(1)} W^{[i]*}_{(2)} = {\Theta}_{i}$, for all $i=1 \dots K$.
\end{proposition}

\noindent The illustration of this idea and extension to nonlinear settings are discussed in Appendix.

\subsection{Positive Transfer across GCN Models}

We apply the theory of \emph{positive transfer} in multi-task learning (MTL) \cite{wu2019understanding} into cross-GCNs learning. If the cross-GCNs training improves over just training the target GCN on one particular graph, we say the source GCNs on other graphs \emph{transfer positively} to the target GCN. \cite{wu2019understanding} proposed a theorem to quantify how many data points are needed to guarantee positive transfer between two ReLU model tasks.
By the above section, it is necessary to limit the capacity of the shared model parameters to enforce knowledge transfer. Here we consider $d = 1$ and $m = m_A = m_B$ to align with the theorem in \cite{wu2019understanding}. We have a theorem for positive transfer between two \emph{one-layer} GCNs that share the first-layer model parameters $W_{(1)}$:
\begin{theorem}
Let $(\hat{A}^{[i]}X^{[i]} \in \mathbb{R}^{n_i \times m}, \vec{\hat{u}}^{[i]} \sqrt{\lambda^{[i]}_1}) \in \mathbb{R}^{n_i}$ denote the data of graph $\mathcal{G}^{[i]}$ ($i=\{A,B\}$). Suppose the output is generated by $ReLU(\hat{A}^{[i]}X^{[i]}~{\theta}^{[i]})$. And suppose $c \geq \mathrm{sin}({\theta}^{[A]}, {\theta}^{[B]}) / \kappa(\hat{A}^{[B]}X^{[B]})$.
Denote by $W^{\star}_{(1)}$ the optimal solution of the one-layer setting of Eq.(\ref{eq:sharew1}): $f(W_{(1)}) = \sum_{i \in \{A, B\}} \mathcal{L} \left( g(X^{[i]}, A^{[i]}, W_{(1)}),~A^{[i]} \right)$. With probability $1-\delta$ over the randomness of $(\hat{A}^{[A]}X^{[A]}, \vec{\hat{u}}^{[A]} \sqrt{\lambda^{[A]}_1})$, when
\begin{equation}
    n_i \geq \mathrm{max}\left( \frac{m \log m}{c^2} (\frac{1}{c^2} + \log m), \frac{{\| \vec{\hat{u}}^{[B]} \sqrt{\lambda^{[B]}_1} \|}^2}{c^2} \right),
\end{equation}
we have that the estimation error is at most:
\begin{equation}
    \mathrm{sin}(W^{\star}_{(1)}, {\theta}^{[A]}) \leq \mathrm{sin}({\theta}^{[A]}, {\theta}^{[B]}) + O(c/\kappa(\hat{A}^{[B]}X^{[B]})).
\end{equation}
\end{theorem}

\noindent \emph{Notations for the above theorem.}
$\kappa(A) = \lambda_{\mathrm{max}}(A) / \lambda_{\mathrm{min}}(A)$ denotes the condition number of $A \in \mathbb{R}^{n \times m}$, where $\lambda_{\mathrm{max}}(A)$ denotes its largest singular value and $\lambda_{\mathrm{min}}(A)$ denotes its min$\{n,m\}$-th largest singular value.
As MTL is effective over ``similar'' tasks \cite{xue2007multi}, the cross-GCNs learning can be effective across similar GCN models. A natural definition of GCN similarity is $\mathrm{cos}(\theta^{[A]}, \theta^{[B]})$: higher means more similar. So $\mathrm{sin}(W^{\star},\theta)=\sqrt{1-\mathrm{cos}^2(W^{\star},\theta)}$ can denote the estimation error. Proof of the theorem can be referred in \cite{wu2019understanding}.

\subsection{Positive Transfer with Network Alignment}

\textbf{Soft regularization}: We transfer knowledge across two GCNs' representation spaces by optimizing Eq.(\ref{eq:alignreg}) where $h$ is specified as Eq.(\ref{eq:softreg}). The optimal solution is
\begin{eqnarray}
    & & U^{[A]\star} = {\left( (1-\beta) \cdot \alpha^{[A]}~I + \beta R^{[A \rightarrow B]} R^{[A \rightarrow B] \top} \right)}^{-1} \cdot \nonumber \\
    & & \cdot {\left( (1-\beta) \cdot \alpha^{[A]}~\hat{U}^{[A]} \sqrt{\Sigma^{[A]}} + \beta~A^{[A,B]} U^{[B]\star} R^{[A \rightarrow B] \top} \right)}.
\end{eqnarray}
So, if $\beta = 0$, $U^{[A]\star} = \hat{U}^{[A]} \sqrt{\Sigma^{[A]}}$ may overfit the graph data $\mathcal{G}^{[A]}$; if $\beta \in (0,1]$, the information transferred from $\mathcal{G}^{[B]}$ via the alignment $A^{[A,B]}$, i.e., $A^{[A,B]} U^{[B]\star} R^{[A \rightarrow B]}$, can have positive impact on link prediction on $\mathcal{G}^{[A]}$.

\vspace{0.05in}
\noindent \textbf{Alignment reconstruction}: Suppose we optimize Eq.(\ref{eq:alignreg} where $h$ is in Eq.(\ref{eq:alignreconstruction}). The optimal solution is
\begin{eqnarray}
    & & U^{[A]\star} = {\left( (1-\beta) \cdot \alpha^{[A]}~\hat{U}^{[A]} \sqrt{\Sigma^{[A]}} + \beta A^{[A,B]} U^{[B]\star} R^{[A,B] \top} \right)} \nonumber \\
    & & \cdot {\left( (1-\beta) \cdot \alpha^{[A]}~I + \beta R^{[A,B]} U^{[B]\star\top} U^{[B]\star} R^{[A,B] \top} \right)}^{-1}.
\end{eqnarray}
So, if $\beta = 0$, $U^{[A]\star} = \hat{U}^{[A]} \sqrt{\Sigma^{[A]}}$ may overfit the graph data $\mathcal{G}^{[A]}$; if $\beta \in (0,1]$, the information transferred from $\mathcal{G}^{[B]}$ via the alignment $A^{[A,B]}$, i.e., $A^{[A,B]} U^{[B]\star} R^{[A,B] \top}$, can have positive impact on link prediction on $\mathcal{G}^{[A]}$.

\begin{table*}[t]
    \centering
    \includegraphics[width=\textwidth]{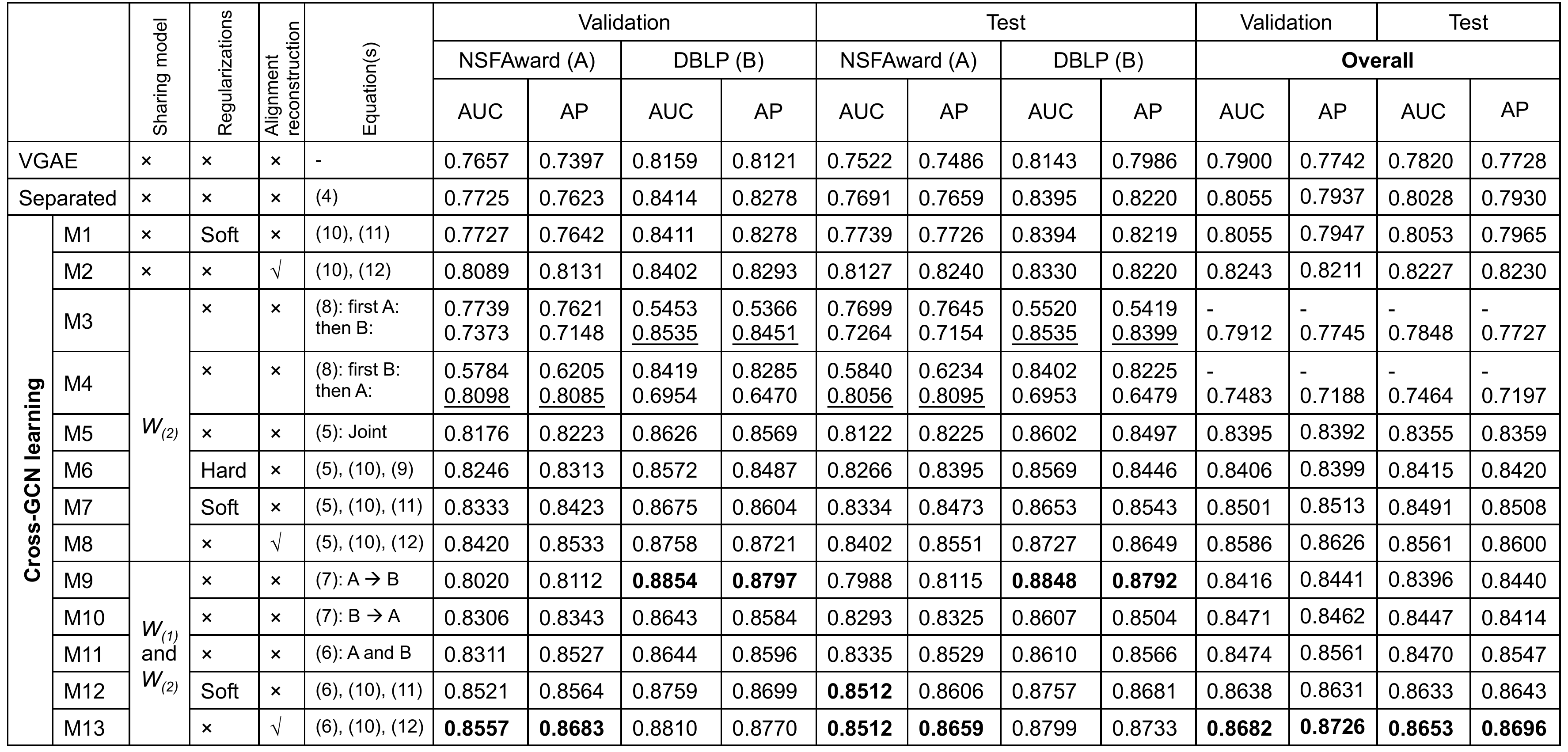}
    \caption{Performance of VGAE \cite{kipf2016variational}, (Separated) GCNs \cite{kipf2017semi}, and 13 cross-GCN learning models on link prediction in NSFAward and DBLP academic graphs ($\mathcal{G}^{[A]}$ and $\mathcal{G}^{[B]}$). The models take various options on model sharing, regularizations, and alignment reconstruction. Equations can be found in Sections 2 and 3. Higher AUC or AP means better performance. We investigated the standard deviation of all the cells: all the values are smaller than 0.0016 and thus omitted for space. Underlined numbers are targeted values in pre-training strategies. Bolded numbers are the highest on the columns.}
    \label{tab:results_link_prediction}
    \vspace{-0.2in}
\end{table*}

\section{Experiments}
\label{sec:experiments}
In this section, I perform cross-network learning on three types of graph learning tasks. For each task, I present datasets (stats can be found in Figure~\ref{fig:datasets}), experimental settings, results, and analysis.

\subsection{Link Prediction}

\subsubsection{Datasets} Below are two collaboration networks I use that can be considered as weighted graphs between researcher nodes.
\begin{itemize}
    \item NSFAward: It is publicly available on the National Science Foundation (NSF) Awards database from 1970 to 2019.\footnote{NSF Award Search: \url{https://www.nsf.gov/awardsearch/download.jsp}} Nodes are investigators of the awards. Two investigator nodes have a link if they are participated in the same award(s) and the weight of the link is the number of their shared award(s).
    \item DBLP: I expand the DBLP-4Area \cite{huang2016meta} from 5,915 authors to 14,387 authors by adding their co-authors in the graph. The weight of the link between two author nodes is the number of their co-authored papers.
\end{itemize}

\begin{figure*}[t]
    \centering
    \includegraphics[width=0.33\textwidth]{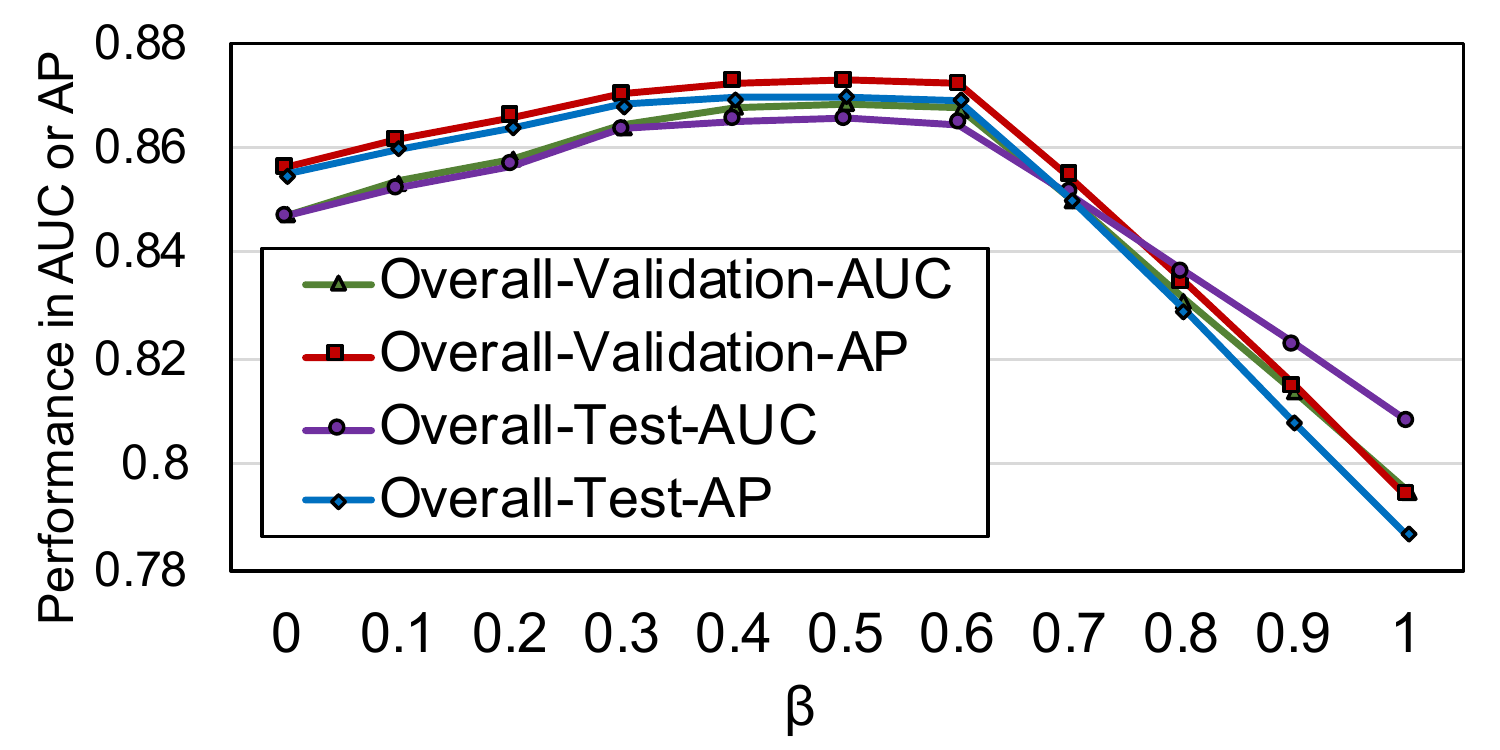}
    \includegraphics[width=0.33\textwidth]{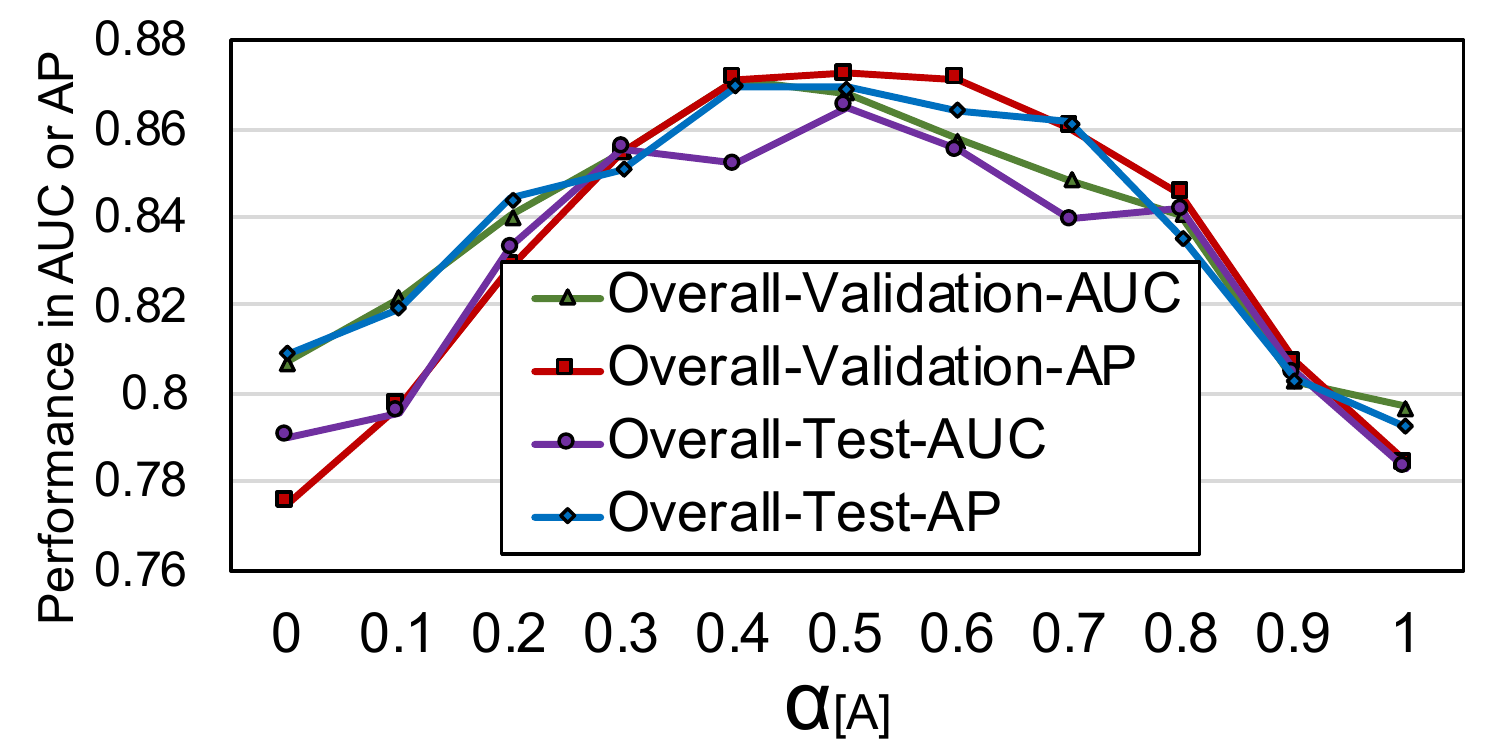}
    \includegraphics[width=0.33\textwidth]{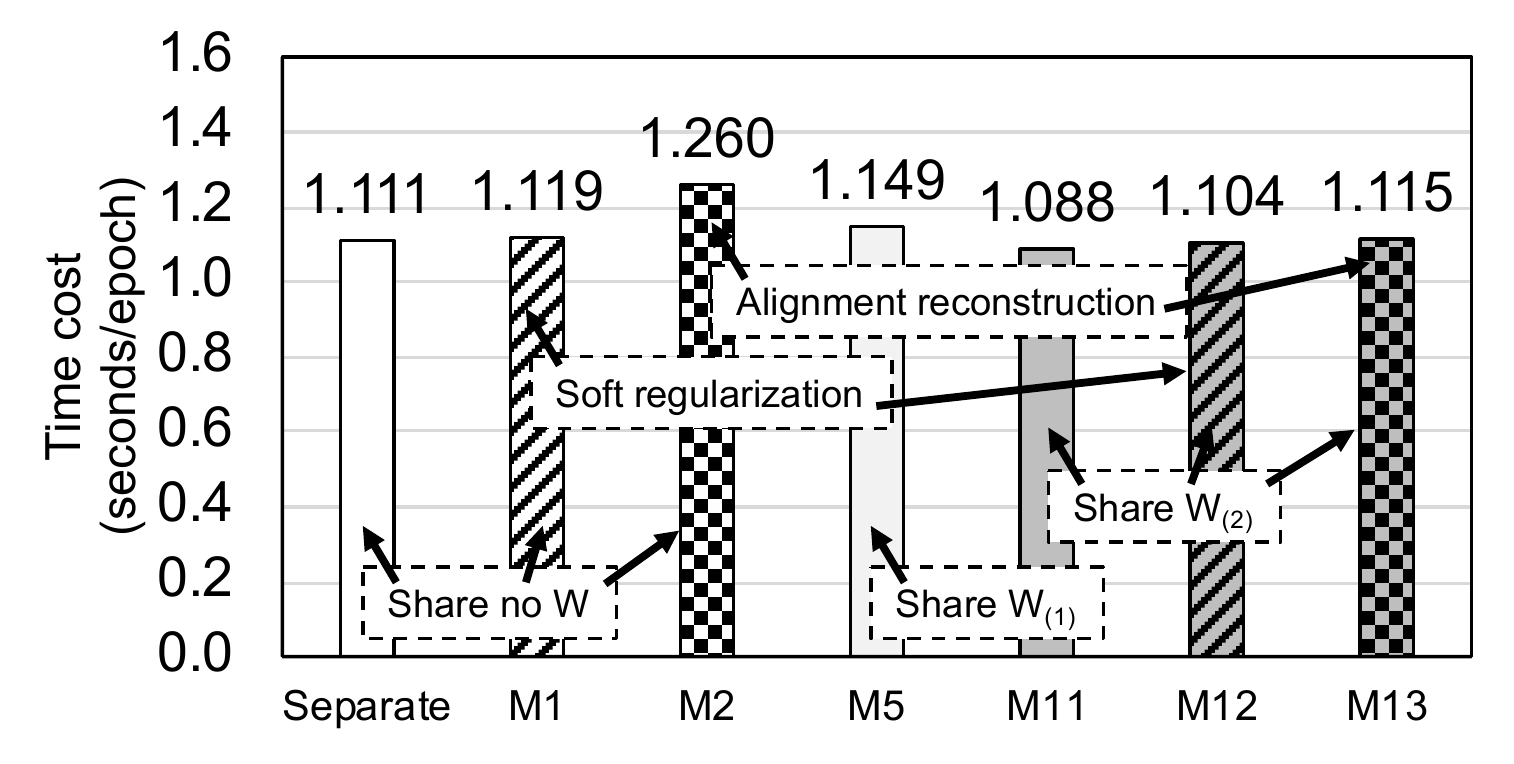}
    \vspace{-0.3in}
    \caption{Sensitivity analysis on M13 (sharing $W_{(1)}$ and $W_{(2)}$, plus alignment reconstruction): (a) on the weight of the alignment reconstruction task $\beta$, (b) on the weight of learning on graph $\mathcal{G}^{[A]}$: $\alpha^{[A]}$. And (c) running time analysis in seconds per epoch.}
    \label{fig:sensitivity_link_prediction}
    \vspace{-0.05in}
\end{figure*}

\subsubsection{Experimental settings} For each graph, I randomly split the observed links into training, validation, and test sets by 8:1:1. Then I randomly select unobserved links as negative examples in the validation and test sets. Positives and negatives are balanced. I use Area Under the ROC Curve (AUC) and Average Precision (AP) as evaluation metrics. I perform the evaluation 10 times, calculate mean value and standard deviation. I also report the ``overall'' performance by calculating the harmonic mean of the AUC/AP on the two datasets. All models are developed by PyTorch and one NVIDIA GeForce RTX 2080 Ti graphic card. I set $d_A = d_B = 64$. I use grid search on the hyperparameters in $\{0.1, 0.2, \dots, 0.9\}$ and find that the best combination is $\alpha^{[A]} = \alpha^{[B]} = \beta = 0.5$.

\subsubsection{Results}
Table~\ref{tab:results_link_prediction} presents the performance of as many as 13 cross-GCN learning models I implemented. Models M3--M8 share $W_{(2)}$ across the NSFAward and DBLP academic graphs. Models M9--M13 share both $W_{(1)}$ and $W_{(2)}$. M6 uses hard regularization; M1, M7, and M12 use soft regularization; and M2, M8, and M13 use alignment reconstruction. I have the following observations:

\vspace{0.05in}
\noindent O1. \emph{Cross-network models perform better than baselines and the best is M13:}
M13 shares $W_{(1)}$ and $W_{(2)}$ between two GCNs, and uses alignment reconstruction to bridge the two representation spaces. In terms of the overall AUC on the two test sets, M13 achieves 0.865, which is significantly higher than VGAE's 0.782 \cite{kipf2016variational} and (Separated) GCNs' 0.803 \cite{kipf2017semi}. The AP is improved from VGAE's 0.773 and GCN's 0.793 to 0.870. Clearly, the knowledge is successfully transferred across the GCN models on two academic graphs. Performance on test sets is worse than but very close to that on validation sets, which indicates no or very little overfitting in the GCN models. Both models need information beyond the corresponding graph data for effective learning and decision-making on the test data.

\vspace{0.05in}
\noindent O2. \emph{Sharing weight matrices is beneficial, and sharing those of both layers is beneficial:} Suppose we do not use the alignment information at all: the only possible bridge across two GCNs is weight matrices $W_{(1)}$ and $W_{(2)}$. M3--M5 share $W_{(2)}$. Compared with separated GCNs, M5 improves the overall AUC on test sets from 0.803 to 0.836. And M9--M11 share both $W_{(1)}$ and $W_{(2)}$. Compared with M5, M11 improves the AUC from 0.836 to 0.847. When soft regularization is adopted, we can compare M7 and M12 with M1; when alignment reconstruction is adopted, we can compare M8 and M13 with M2. The conclusions are the same.

\begin{figure}[t]
    \centering
    \includegraphics[width=0.48\textwidth]{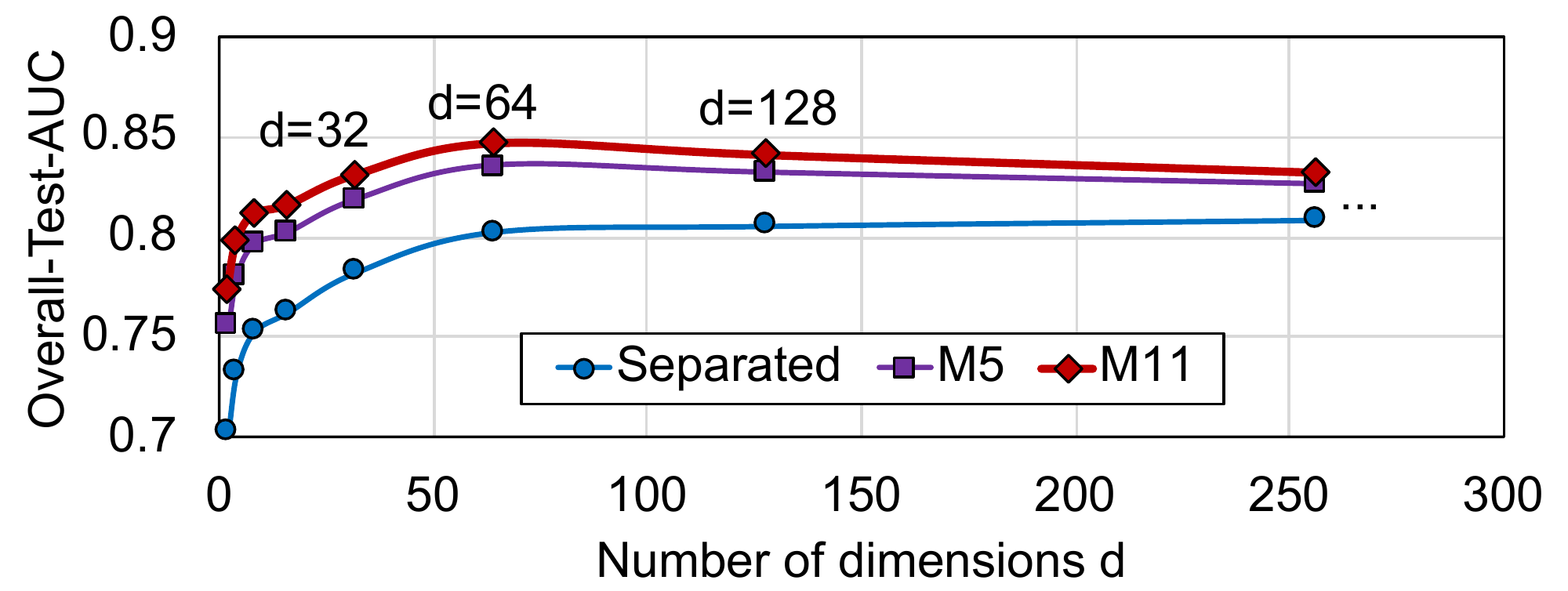}
    \vspace{-0.3in}
    \caption{Empirical analysis on the choice of $d$.}
    \label{fig:choice_d}
    \vspace{-0.2in}
\end{figure}

\vspace{0.05in}
\noindent O3. \emph{If $W_{(2)}$ is shared, joint training is slightly better than pre-training:} M3 pre-trains the model with $\mathcal{G}^{[A]}$ and then fine-tunes on $\mathcal{G}^{[B]}$; M4 switches the use of the graphs. So we focus on M3's performance on $\mathcal{G}^{[B]}$ and M4's on $\mathcal{G}^{[A]}$. Compared with separated GCNs, M3 improves AUC on the test set from 0.840 to 0.854; M4 improves it from 0.769 to 0.806. However, with a joint training strategy, M5 achieves an AUC on $\mathcal{G}^{[B]}$ of 0.860 over M3's 0.854 and achieves an AUC on $\mathcal{G}^{[A]}$ of 0.812 over M4's 0.806.

\vspace{0.05in}
\noindent O4. \emph{If $W_{(1)}$ is shared, learning linear transformation matrices on both graphs is slightly better than learning a matrix on one of them:} M11 learns $Q^{[A]} \in \mathbb{R}^{m_A \times \hat{m}}$ and $Q^{[B]} \in \mathbb{R}^{m_B \times \hat{m}}$. M9 learns $Q^{[A \rightarrow B]}$ only, and M10 learns $Q^{[B \rightarrow A]}$ only. In terms of the overall AUC on the two test sets, compared with M5 that shares $W_{(2)}$ only, M9 improves AUC from M5's 0.836 to 0.840, M10 achieves an AUC of 0.845, and M11 achieves an AUC of 0.847.

\vspace{0.05in}
\noindent O5. \emph{Regularizations, hard or soft, are beneficial; and alignment reconstruction is better:} (a) M6 adds hard regularization to M5. M6 improves the overall AUC on the two test sets from M5's 0.836 to 0.842. (b) M1, M7, and M12 add soft regularization to separated GCNs, M5, and M11, respectively. M1 improves the AUC from 0.803 to 0.805, M7 improves from 0.836 to 0.849, and M13 improves from 0.847 to 0.863. With more model parameters (e.g., weight matrices) shared, the improvements by regularizations become bigger. (c) M8 and M13 add alignment reconstruction to M5 and M11, respectively. M8 improves the AUC from 0.836 to 0.856. M13 improves from 0.847 to 0.865. The AUC scores are quite significantly improved by alignment reconstruction, slightly higher than regularizations.

\subsubsection{Hyperparameter sensitivity} In Figure~\ref{fig:sensitivity_link_prediction}(a) and~(b) we investigate the effect of hyperparameters $\beta$ and $\alpha^{[A]}$ on the performance of model M13. When $\beta=0$, M13 is equivalent with M11. With $\beta$ increasing, the performance becomes better by learning to reconstruct the network alignment. It is the best when $\beta \in [0.4,0.6]$. Then a too big $\beta$ will overfit the model on the alignment reconstruction, so the AUC/AP decreases quickly when $\beta$ goes from 0.6 to 1.
Given $\beta=0.5$, a too big or too small $\alpha^{[A]}$ will overfit one of the graph data, $\mathcal{G}^{[A]}$ or $\mathcal{G}^{[B]}$. When $\alpha^{[A]}$ is around 0.5, the performance is the best. So, all the optimization terms are important.

\subsubsection{Time complexity} Figure~\ref{fig:sensitivity_link_prediction}(c) presents the time cost per epoch of separated GCNs, M1, M2, M5, M11--M13. The models that share both $W_{(1)}$ and $W_{(2)}$ (M11), share $W_{(2)}$ only (M5), and share neither of them (Separated) do not show significant difference of time cost. Adopting soft regularizations (M1 vs Separated, M12 vs M11) increases the time cost slightly. And adopting alignment reconstruction (M2 vs M1, M13 vs M12) take slightly more time. The time complexity remains at the same level in the proposed models.

\subsubsection{On the choice of $d$}
Figure~\ref{fig:choice_d} presents the overall AUC on test sets of Separated GCNs, M5 (sharing $W_{(2)}$), and M11 (sharing $W_{(1)}$ and $W_{(2)}$), when the number of dimensions $d$ is $\{2, 2^2, \dots, 2^8\}$.
When $d$ becomes bigger, the improvements by model parameters sharing become smaller. As presented in theoretical analysis, small $d$ would force the knowledge transfer across the GCN models; when $d$ is too big, there might be no transfer. We choose $d=64$ as the default number of dimensions of the latent spaces.

\subsection{Relation Classification}

\subsubsection{Datasets} Below are four knowledge graphs I use that can be considered as relational graphs between entity nodes.
\begin{itemize}
    \item FB15K: It is a subset of the relational database Freebase \cite{schlichtkrull2018modeling}.
    \item WN18: It is a subset of WordNet containing lexical relations.
    \item YAGO: The open source knowledge base developed by MPI.
    \item DBpedia500: A subset of structured content in the Wikipedia.
\end{itemize}
I will investigate whether incorporating partially-aligned graphs YAGO and DBPedia500 can improve the performance of relation classification on FB15K and WN18, respectively. Entities are aligned if they have the same surface name. Resolving the inaccuracy of entity alignment is important but out of this work's scope.

\begin{table}[t]
    \centering
    \includegraphics[width=0.48\textwidth]{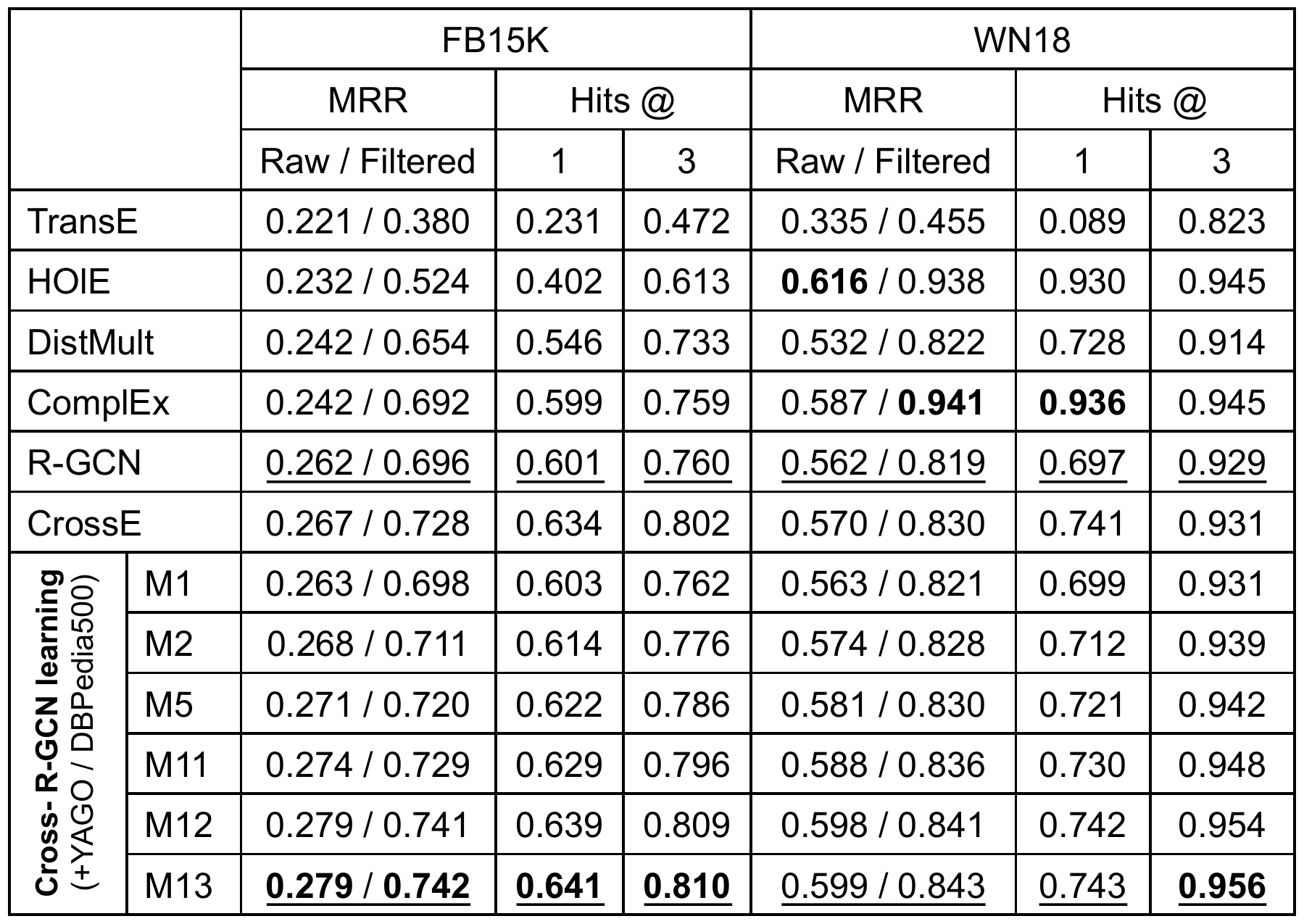}
    \caption{Performance of existing methods (e.g., RGCN \cite{schlichtkrull2018modeling}) and cross-R-GCN learning models on relation classification.}
    \label{tab:results_relation_classification}
    \vspace{-0.3in}
\end{table}

\subsubsection{Experimental settings} Both FB15K and WN18 have standard splits for training, validation, and test. I provide results using two common evaluation metrics: mean reciprocal rank (MRR) and Hits at $k$ (Hits@$k$). I report both raw and filtered MRR (with filtered MRR typically considered more reliable) and filtered Hits@1 and Hits@3, following TransE \cite{bordes2013translating}. Other baseline methods include:
\begin{itemize}
    \item HOlE \cite{nickel2016holographic}: it replaces vector-matrix product with circular correlation to compute knowledge graph embeddings.
    \item DistMult \cite{yang2014embedding}: a neural embedding approach that learns representations of entities and relations.
    \item ComplEx \cite{trouillon2016complex}: it models asymmetric relations by generalizing DistMult to the complex domain. 
    \item R-GCN \cite{schlichtkrull2018modeling}: it is the first work of relation graph convolutional network, considered the state of the art. And I use it as the basic GCN model for cross-GCN learning.
    \item CrossE \cite{zhang2019interaction}: its ``interaction embeddings" are learned from both knowledge graph connections and link explanations.
\end{itemize}
I use the Adam optimizer with a learning rate of 0.01. Implementation details can be found in Appendix.

\subsubsection{Results} Table~\ref{tab:results_relation_classification} presents the results of the baselines and six cross-GCN learning models (M1, M2, M5, M11--M13). Model M5 shares $W_{(2)}$ across two knowledge graphs. M11--M13 share both $W_{(1)}$ and $W_{(2)}$. M1 and M12 use soft regularization; and M2 and M13 use alignment reconstruction. Our main observation is that \emph{cross-network models perform better than baselines and the best model is M13}.
With the large-scale YAGO, M13 improves the filtered MRR from 0.696 to 0.742 and the filtered Hits@1 from 0.601 to 0.641 for relation classification on FB15K, compared to a sole R-GCN model.
With DBPedia500 which is bigger than YAGO, M13 improves the filtered MRR from 0.819 to 0.843 and the filtered Hits@1 from 0.697 to 0.743 on WN18, compared to R-GCN.
Both M12 and M13 that share two-layers parameters and alignment information achieve higher performance than the best baseline CrossE on all the metrics.

\section{Related Work}
\label{sec:related}
\subsection{Graph Neural Networks}

Graph neural network (GNN) is a type of neural network that operates on graph structure to capture the dependence of graphs via message passing between the nodes of graphs \cite{zhang2020deep,wu2020comprehensive}. Kipf \emph{et al.} proposed graph convolutional networks that use a spectral graph convolution to capture the spatial dependency \cite{kipf2017semi}. Velivckovic \emph{et al.} considered the interaction between nodes in the graph convolution measured as graph attention weight \cite{velivckovic2017graph}. Hamilton \emph{et al.} implemented the graph convolution as an efficient aggregation and sampling algorithm for inductive node representation learning \cite{hamilton2017inductive}.

Recently, GNN techniques have been applied for learning representations on relational graphs, specifically knowledge graphs \cite{sadeghian2019drum,zhang2019quaternion,ren2020beta}. Relational GNNs perform significantly better than conventional knowledge graph embeddings \cite{schlichtkrull2018modeling}.
Most techniques suffer from the issue of data sparsity and our proposed methods address it with knowledge transfer across networks.

In this paper, we discussed weighted graph, relational graph, and bipartite graphs; however, heterogeneous graphs are ubiquitous and have unique essential challenges for modeling and learning \cite{wang2020heterogeneous,luo2020dynamic,liu2020heterogeneous,wang2019attentional,wang2019kgat}.
We have been seeing heterogeneous GNNs \cite{dong2017metapath2vec,liu2018heterogeneous,zhang2019heterogeneous,fu2020magnn} and heterogeneous attention networks for graphs \cite{wang2019heterogeneous,yang2020heterogeneous,hu2020heterogeneous,yao2020heterogeneous,hu2020gpt}.
Extending our proposed cross-network learning to heterogeneous graphs and enabling inductive learning in the framework are very important. These two points are out of the scope of our work but worth being considered as future directions.

\subsection{Network Alignment and Modeling}

The alignment of nodes between two or more than two graphs has been created and/or utilized in at least three categories of methods \cite{jiang2012social,jiang2015social,vijayan2017multiple,chen2017community,qin2020g,gu2020data}.
First, Kong \emph{et al.} inferred anchor links between two social networks by matching the subgraphs surrounding the linked nodes \cite{kong2013inferring}. Emmert \emph{et al.} wrote a comprehensive survey on the algorithms of graph matching \cite{emmert2016fifty}.
Second, Zhang \emph{et al.} used graph pattern mining algorithms for node alignment on incomplete networks \cite{zhang2017ineat} and multi-level network \cite{zhang2019multilevel}. Nassar \emph{et al.} leveraged the multimodal information for network alignment \cite{nassar2017multimodal}. The third line of work learned node representations and trained the supervised models with node pairing labels (if available) \cite{du2019joint,ye2019vectorized,nassar2018low}. A matrix factorization-based model has demonstrated that cross-network or cross-platform modeling can alleviate the data sparsity on a single platform \cite{jiang2016little}.



\section{Conclusions}
\label{sec:conclusions}
In this paper, I proposed partially aligned GCNs that jointly learn node representations across graphs.
I provided multiple methods as well as theoretical analysis to positively transfer knowledge across the set of partially aligned nodes.
Extensive experiments on real-world knowledge graphs and collaboration networks show the superior performance of the proposed models on relation classification and link prediction.

\section*{Acknowledgment}
This research was supported by National Science Foundation award IIS-1849816.

\balance
\bibliographystyle{ACM-Reference-Format}
\bibliography{main}

\newpage
\appendix

\section{Additional Information for Theoretical Analysis}

\subsection{Choices of the Number of Dimensions $d$}

The total capacities (i.e., sum of ranks of adjacency matrices) can be implemented as $\sum^{K}_{i=1} \mathrm{argmax}_{p}~\lambda_{p}(A^{[i]}) \geq \epsilon$, the sum of the minimum positions of no-smaller-than-$\epsilon$ singular value. To illustrate the idea, suppose $\{W^{[i]}_{(2)}\}$ are shared across GCNs, which means $W_{(2)}=W^{[1]}_{(2)}=\dots=W^{[K]}_{(2)}$. As long as the shared $W^{*}_{(2)}$ contains ${{\Theta}_{i}}|^{K}_{i=1}$ in its column span, there exits $W^{[i]*}_{(1)}$ such that $W^{[i]*}_{(1)} W^{*}_{(2)} = {\Theta}_{i}$, which is optimal for Eq.(\ref{eq:losslinear}) with minimum error. But this means no transfer across any GCNs. This can hurt generalization if a GCN has limited data $\mathcal{G}^{[i]}$, in which its single-GCN solution overfits training data, whereas the cross-network solution can leverage other graphs' data to improve generalization.

\vspace{0.05in}
\noindent \textbf{Extension to the ReLU/softmax setting.} If the capacity of the shared model parameters (i.e., the number of dimensions $d$) is larger than the total capacities, then we can share both ${\{W^{[i]}_{(1)}\}}|^{K}_{i=1}$ and ${\{W^{[i]}_{(2)}\}}|^{K}_{i=1}$. This remains an optimal solution to the joint training problem in the ReLU/softmax setting. Furthermore, there is \emph{no transfer} between any GCNs through the shared model parameters.

\subsection{Positive Transfer with Network Alignment}

\textbf{Soft regularization}: The loss function can be re-written as:
\begin{eqnarray}
    f & = & (1-\beta) \cdot \alpha^{[A]}~{\| U^{[A]} - \hat{U}^{[A]} \sqrt{\Sigma^{[A]}} \|}^2_\mathrm{F} \nonumber \\
    & & +~(1-\beta) \cdot \alpha^{[B]}~{\| U^{[B]} - \hat{U}^{[B]} \sqrt{\Sigma^{[B]}} \|}^2_\mathrm{F} \nonumber \\
    & & +~\beta~{\| U^{[A]} R^{[A \rightarrow B]} - A^{[A,B]} U^{[B]} \|}^2_\mathrm{F}.
\end{eqnarray}
The gradient is
\begin{eqnarray}
    & & \frac{\mathrm{d}~f}{\mathrm{d}~U^{[A]}} = 2~(1-\beta) \cdot \alpha^{[A]}~(U^{[A]} - \hat{U}^{[A]} \sqrt{\Sigma^{[A]}}) \nonumber \\
    & & + 2~\beta~(U^{[A]} R^{[A \rightarrow B]} - A^{[A,B]} U^{[B]}) R^{[A \rightarrow B]~\top}.
\end{eqnarray}

\vspace{0.05in}
\noindent \textbf{Alignment reconstruction}: The loss function is:
\begin{eqnarray}
    f & = & (1-\beta) \cdot \alpha^{[A]}~{\| U^{[A]} - \hat{U}^{[A]} \sqrt{\Sigma^{[A]}} \|}^2_\mathrm{F} \nonumber \\
    & & +~(1-\beta) \cdot \alpha^{[B]}~{\| U^{[B]} - \hat{U}^{[B]} \sqrt{\Sigma^{[B]}} \|}^2_\mathrm{F} \nonumber \\
    & & +~\beta~{\| U^{[A]} R^{[A,B]} U^{[B]\top} - A^{[A,B]} \|}^2_\mathrm{F}.
\end{eqnarray}
The gradient is
\begin{eqnarray}
    & & \frac{\mathrm{d}~f}{\mathrm{d}~U^{[A]}} = 2~(1-\beta) \cdot \alpha^{[A]}~(U^{[A]} - \hat{U}^{[A]} \sqrt{\Sigma^{[A]}}) \nonumber \\
    & & + 2~\beta~(U^{[A]} R^{[A,B]} U^{[B]\top} - A^{[A,B]}) U^{[B]} R^{[A,B]~\top}.
\end{eqnarray}

\end{document}